\documentclass{article}
\usepackage{amsmath,amsfonts}
\usepackage{algorithmic}
\usepackage{algorithm}
\usepackage{array}
\usepackage{booktabs}
\usepackage{threeparttable}
\usepackage{multirow}
\usepackage{multicol}
\usepackage{diagbox}
\usepackage[caption=false,font=normalsize,labelfont=sf,textfont=sf]{subfig}
\usepackage{textcomp}
\usepackage{stfloats}
\usepackage{url}
\usepackage{verbatim}
\usepackage{graphicx}
\usepackage{cite}

\usepackage{soul}
\usepackage{booktabs}
\usepackage{xcolor}
\usepackage{bbding}
\usepackage[misc]{ifsym}

\usepackage[breaklinks,colorlinks]{hyperref}
\usepackage[capitalize]{cleveref}
\crefname{section}{Sec.}{Secs.}
\Crefname{section}{Section}{Sections}
\Crefname{table}{Table}{Tables}
\crefname{table}{Tab.}{Tabs.}
\usepackage{spconf,amsmath,epsfig}

\let\OLDthebibliography\thebibliography
\renewcommand\thebibliography[1]{
  \OLDthebibliography{#1}
  \setlength{\parskip}{0pt}
  \setlength{\itemsep}{0pt plus 0.3ex}
}

\begin{document}\sloppy

\def\x{{\mathbf x}}
\def\L{{\cal L}}

\title{SDFReg: Learning Signed Distance Functions for Point Cloud Registration}
%
\name{Leida Zhang, Zhengda Lu, Kai Liu, Yiqun Wang\textsuperscript{\Envelope}}
\address{}

\maketitle
\begin{abstract}
   Learning-based point cloud registration methods can handle clean point clouds well, while it is still challenging to generalize to noisy, partial, and density-varying point clouds. To this end, we propose a novel point cloud registration framework for these imperfect point clouds. By introducing a neural implicit representation, we replace the problem of rigid registration between point clouds with a registration problem between the point cloud and the neural implicit function. We then propose to alternately optimize the implicit function and the registration between the implicit function and point cloud. In this way, point cloud registration can be performed in a coarse-to-fine manner. By fully capitalizing on the capabilities of the neural implicit function without computing point correspondences, our method showcases remarkable robustness in the face of challenges such as noise, incompleteness, and density changes of point clouds. 

\end{abstract}


\section{Introduction}
\label{sec:intro}
Point cloud registration is one of the crucial research topics of computer vision and computer graphics and has important applications in the fields of 3D reconstruction, autonomous driving, and intelligent robotics. 
Its goal is to solve the transformation matrix of two point clouds with different poses and align them at the same coordinate system. 
Most existing methods for point cloud registration can be divided into correspondence-based and direct registration methods.

Correspondence-based registration methods \cite{ICP,PLICP,DCP,3D-3D} first extract corresponding points between two point clouds and then solve the transformation. These methods can be further categorized into two classes. 
The first class\cite{ICP,PLICP} minimizes geometric projection errors by adopting iterative optimization strategies of correspondence search and transformation estimation.
The second class\cite{DCP,3D-3D} leverages deep neural networks to search correspondence by learnable features. Then, the transformation matrix is solved without iterations. However, the correspondence search is easily affected by noise, incompleteness, and density changes in point clouds, making the overall scheme less robust.

\begin{figure}[t]
  \centering
   \includegraphics[width=0.85\linewidth]{"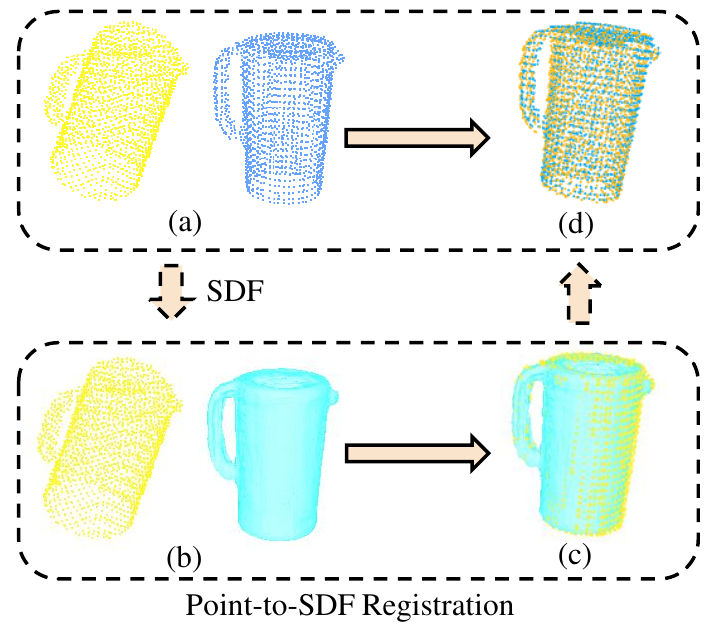"}
    \vspace{-1.0em}
   \caption{Our method makes full use of a signed distance function (SDF) for point cloud registration: (a) Initial position of source (yellow) and target (blue) point sets; (b) Then the target set is represented by an SDF; (c) The registration result of the source set and the SDF is obtained; (d) The output point cloud registration results of the target set (blue) and the transformed source set (yellow).}
   \label{fig1}
   
\end{figure}

Different from the correspondence-based registration method, direct methods estimate transformation directly without computing corresponding points. Some traditional direct methods \cite{correspondencefree,richer} use implicit polynomials to fit the target point cloud. 
The pose is finally estimated by optimizing the distance error between the source point cloud and the implicit polynomials. 
However, these registration methods are only verified on smooth surfaces due to the fitting ability of polynomials and cannot register fine objects or complex scenes.
Recently, more direct registration methods\cite{PointNetLK,FMR,OMNet,REGTR, ifr, DeepGMR} have been proposed to solve complex point cloud registration problems, which model the transformation estimator as a deep neural network to optimize the pose transformation. These methods transform the registration problem from minimizing point-point projection error to minimizing feature difference. While these methods excel in handling clean data, the influence of outliers within point clouds introduces a distortion in the distance metric based on feature differences, resulting in a suboptimal registration quality.

To this end, we propose a novel framework for point cloud registration. As shown in \cref{fig1}, we introduce a neural implicit function into the pipeline of point cloud registration. Our method represents the target point cloud as a neural implicit surface, i.e. learnable signed distance function (SDF). Then the difference between two point clouds can be transformed into the distance metric between the source point cloud and the neural implicit surface. 
By optimizing the distance metric, the pose of two point clouds can be calculated. Our method transforms the rigid registration problem between two point clouds into point clouds and implicit functions. 
Subsequently, we iteratively refine the optimization of both the SDF and the registration alignment between the SDF and the point cloud, which enables a coarse-to-fine point cloud registration process. 
By transforming the rigid registration problem between two point clouds into a problem involving both point cloud and neural implicit surface, our method can effectively handle noise, incompleteness, and density variations within the point clouds.
For noise and partial point clouds, our method performs better compared with all competitors (ICP\cite{ICP}, PointNetLK\cite{PointNetLK}, DCP\cite{DCP}, FMR\cite{FMR}, DeepGMR\cite{DeepGMR}, and IFR\cite{ifr}). For point clouds with a significant density difference, DeepGMR\cite{DeepGMR} encounters challenges in normal execution. In contrast, our method not only sustains excellent performance but also attains results on par with the IFR\cite{ifr}. Our method exhibits commendable robustness in dealing with imperfect point cloud data.

The main contributions of our paper are as follows:
\begin{itemize}
\item [(1)] 
We propose SDFReg, a novel point cloud registration framework that fully leverages the capabilities of the neural implicit function, eliminating the necessity to search for corresponding points.
\vspace{-0.8em}
\item [(2)] 
We present a coarse-to-fine learning strategy to alternately optimize the implicit function fitting and the registration, which further improves the registration quality. 
\vspace{-0.8em}
\item [(3)] 
Our proposed method is robust to the defects of point clouds and can be generalized to different categories.
\end{itemize}

\vspace{-1.0em}
\begin{figure}[ht]
\hspace{-3.8mm}
    \centering
    \includegraphics[width=0.5\textwidth]{"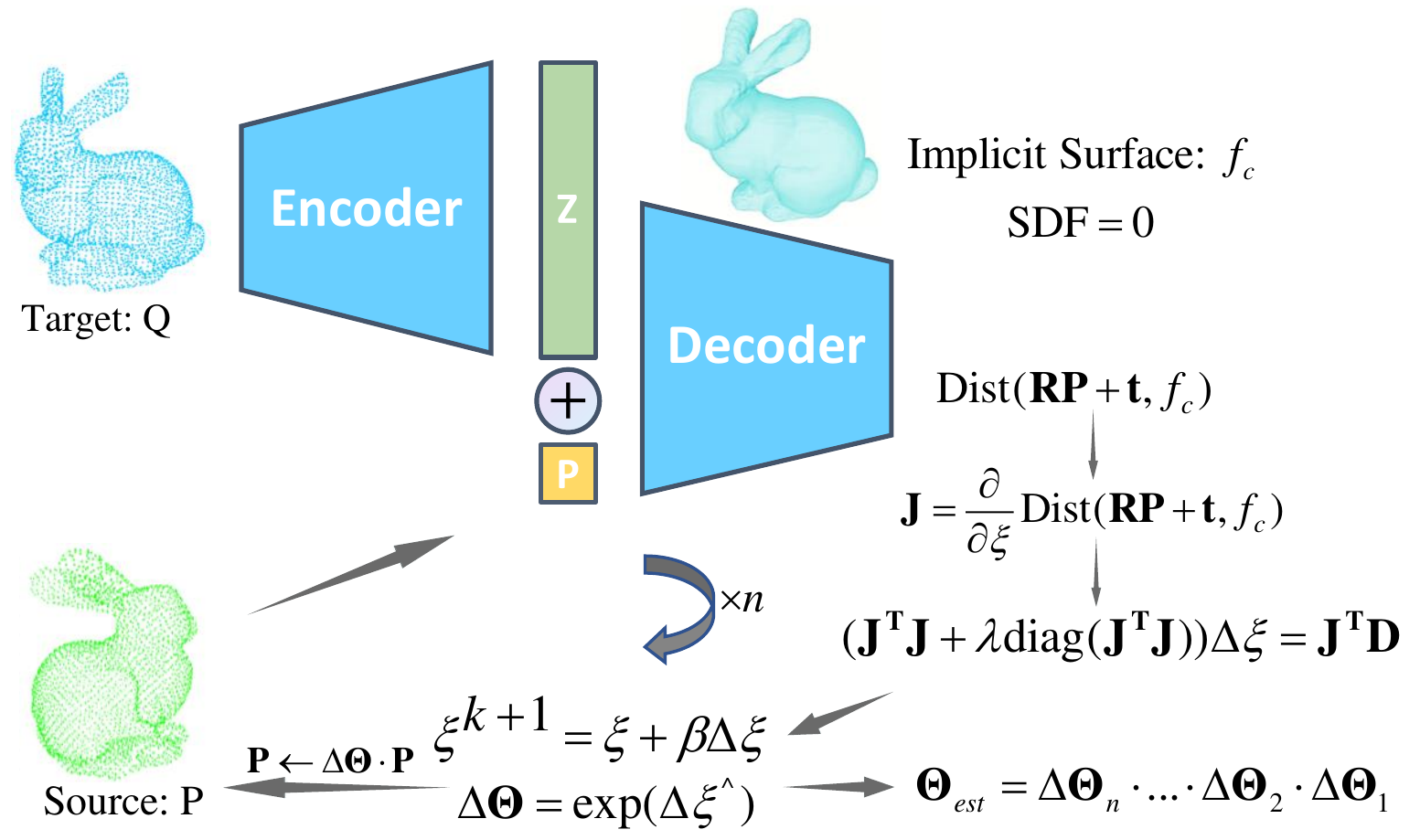"}
    \vspace{-1.0em}
    \caption{The structure of our framework consists of encoder and decoder parts. First, the encoder extracts the features $z$ for the target input point cloud. The encoder section only needs to be done once. Then the coordinates of the source point cloud and the feature $z$ are concatenated and input to the decoder. Furthermore, the distance between the source point and the implicit surface $f_c$ will be decoded as the value of SDF on the source point, $\operatorname{Dist}\left(\mathbf{RP}+\mathbf{t}, f_{c}\right)=\operatorname{SDF}\left(\mathbf{RP}+\mathbf{t}\right)$. Meanwhile, the $\mathbf{J}$ is the Jacobian matrix of $\operatorname{Dist}\left(\mathbf{RP}+\mathbf{t}, f_{c}\right)$ with respect to transformation parameters $\mathbf{\xi}$. Since the distance error is in the non-linear least squares form, it can be optimized by the Levenberg-Marquardt algorithm to estimate transformation increment $\mathbf{\Delta\xi}$ and iteratively update the transformation parameters $\mathbf{\Delta\Theta}$.}

    \label{fig2}
\end{figure}

\section{Method}
\label{sec:3}

As illustrated in \cref{fig2}, we propose SDFReg, which consists of two main components for point cloud registration. 
In the first component, we represent the target point cloud as a neural implicit surface, which is defined as a learnable SDF. 
We then replace the distance computation between the source point cloud and the target point cloud with the distance between the neural implicit surface and the target to determine the registration error. 
In the second component, we optimize the registration error using the SDF through its gradient information and find the optimal rigid parameters that minimize the proposed losses. 
Within this framework, we propose an alternate optimization of these two components in a coarse-to-fine manner.

\subsection{Problem Statement} 
\label{ssec:3.2}
3D rigid point cloud registration refers to finding a rigid transformation $\mathbf{\Theta} \in \mathbb{R}^{4 \times 4}$ to align the source point cloud $\mathbf{P}=\left\{\mathbf{p}_{i} \in \mathbb{R}^{3} \mid i=1,2, \ldots, N\right\}$  and the target point cloud $\textbf{Q}=\left\{\mathbf{q}_{i} \in \mathbb{R}^{3} \mid i=1,2, \ldots, M\right\}$. 
The pose transformation $\mathbf{\Theta}$ describes the motion of a rigid body during registration which can be expressed as follows, 
    \begin{equation}
        \mathbf{\Theta} =\left(\begin{array}{ll}
        \mathbf{R} & \mathbf{t} \\
        \mathbf{0}^{\mathbf{T}} & {1}
        \end{array}\right) \in \mathrm{SE}(3)
        \label{equation 3}
    \end{equation}
where $\mathbf{R} \in \operatorname{SO(3)}$ is the rotation matrix, $\mathbf{t} \in \mathbb{R}^{3}$ is the translation vector, N and M respectively represent the number of points in $\mathbf{P}$ and $\textbf{Q}$. The rigid motion is represented by $\xi^{\wedge} \in \mathfrak{s e}(3)$, where the twists are parametrized by a six-dimensional vector in the form of $\xi=\left[\phi_{1}, \phi_{2}, \phi_{3}, \rho_{1}, \rho_{2}, \rho_{3}\right]^{T} \in \mathbb{R}^{6}$.

The rigid point cloud registration problem can be described as solving the optimal parameter to move the source point cloud $\mathbf{P} \in \mathbb{R}^{N \times 3}$ closest to the target point cloud $\textbf{Q} \in \mathbb{R}^{M \times 3}$ :
    \begin{equation}
        \hat{\mathbf{\Theta}}=\underset{\mathbf{R} \in SO(3),\mathbf{t \in \mathbb{R}^{3}} }{\operatorname{argmin}}\operatorname{||}\text{T}(\mathbf{R} \mathbf{P}+\textbf{t})_{,} \text{T}(\textbf{Q})\operatorname{||}_{2}^{2}
        \label{equation 36}
    \end{equation}
where $\mathbf{T}$ is used differently in different types of registration methods. In the direct registration method, $\mathbf{T}$ is a function used to extract features. In correspondence-based registration methods, $\mathbf{T}$  represents the transformation based on the sampling with the same number of points and the order.

\subsection{Registration Strategy using SDF} 
\label{ssec:3.3}

When matching imperfect point clouds, such as those affected by noise, incompleteness, and density changes, these imperfections can impact the correspondence search in correspondence-based registration methods and the feature extraction and representation in direct registration methods.

In order to solve the problems encountered by the above two types of registration methods, we propose to transform the rigid registration between two point clouds into a registration problem involving the point cloud and the neural implicit surface where the implicit surface is introduced to represent point cloud $\textbf{Q}$ as $f_{c}$. As a result, \cref{equation 36} can be replaced with:
    \begin{equation}
        \hat{\mathbf{\Theta}}=\underset{\mathbf{R} \in SO(3),\mathbf{t \in \mathbb{R}^{3}} }{\operatorname{argmin}}\left(\sum_{i=1}^{N} \operatorname{Dist}{ }^{2}\left(\mathbf{R} \mathbf{p}_{i}+\mathbf{t}_{,} f_{c}\right)\right)
        \label{equation 6}
    \end{equation}
where $\operatorname{Dist}$ represents the orthogonal distance function between a point and the implicit surface $f_{c}$.

In this work, we use the signed distance function (SDF) to represent the target point cloud $\textbf{Q}$. For a given 3D query point $\mathbf{x} \in \mathbb{R}^{3}$, the SDF is the distance from the point to the nearest point on the surface. Its sign represents whether the point is inside (negative) or outside (positive) of the surface.

    \begin{equation}
        \mathbf{x} \rightarrow \operatorname{SDF}(\mathbf{x} \mid \textbf{Q})=s, s \in \mathbb{R}
        \label{equation 1}
    \end{equation}
    
The distance between the point $\mathbf{R} \mathbf{p}_{i}+\mathbf{t}$ and the implicit surface $f_{c}$ can then be represented by the value of SDF.
    \begin{equation}
        \operatorname{Dist}\left(\mathbf{R} \mathbf{p}_{i}+\mathbf{t}, f_{c}\right)=\operatorname{SDF}\left(\mathbf{R} \mathbf{p}_{i}+\mathbf{t}\right)
        \label{equation 7}
    \end{equation}

The distance metric provides a correspondence-free formulation for the registration problem, which is directly related to the rigid parameter. In order to solve \cref{equation 6} by the gradient-based optimization algorithm, the Jacobian matrix $\mathbf{J}$ of the distance metric is required. In the 3D case, the Jacobian matrix $\mathbf{J}$ consisting of $N \times 6$ elements which are calculated by taking partial derivatives, and the value of $i$-th row, the $j$-th column is given.

    \begin{equation}
        \begin{aligned}
            \operatorname{J}(i, j) &=\frac{\partial}{\partial \xi_{j}} \operatorname{SDF}\left(\mathbf{R} \mathbf{p}_{i}+\mathbf{t}\right)
            \end{aligned}
        \label{equation 9}
    \end{equation}
By applying the chain rule, the final Jacobian matrix can be obtained as follows.
    \begin{equation}
        \operatorname{J}(i, j)=\frac{\partial}{\partial \xi_{j}}\left(\mathbf{R} \mathbf{p}_{i}+\mathbf{t}\right) \cdot \nabla \operatorname{SDF}\left(\mathbf{R} \mathbf{p}_{i}+\mathbf{t}\right)
        \label{equation 10}
    \end{equation}
The implicit function represented by the SDF fitted by the neural network is smooth, the gradient $\nabla \operatorname{SDF}\left(\mathbf{R}\mathbf{p}_{i}+\mathbf{t}\right) \in \mathbb{R}^{3}$ of the SDF can be obtained through the backpropagation of the network, and $\partial\left(\mathbf{R} \boldsymbol{\mathbf{p}}_{i}+\mathbf{t}\right) / \partial \xi_{j} \in \mathbb{R}^{3}$ can be calculated following \cite{richer}.

After estimating the distance metric and calculating its Jacobian matrix via \cref{equation 9} and \cref{equation 10}, the LMA can be performed to refine the iterative increment $\Delta\xi$ of the rigid body transformation parameters:
    \begin{equation}
        \left(\mathbf{J}^{\mathbf{T}} \mathbf{J}+\lambda \operatorname{diag}\left(\mathbf{J}^{\mathbf{T}} \mathbf{J}\right)\right) \Delta \xi=\mathbf{J}^{\mathbf{T}} \mathbf{D}
        \label{equation 11}
    \end{equation}
where $\mathbf{D} \in \mathbb{R}^{N}$ is a column vector containing vector $\operatorname{Dist}\left(\mathbf{R} \mathbf{p}_{i}+\mathbf{t}, f_{c}\right)$; $\lambda$ is the damping parameter in LMA; the vector $\Delta\xi$ represents the refinement vector for the rigid parameters.

The result $\Delta\xi$ produced by \cref{equation 11} is mapped into its corresponding transformation matrix in $\operatorname{SE(3)}$ and used to update the source point cloud:
    \begin{equation}
        \Delta \boldsymbol{\Theta}=\exp \left(\Delta \xi^{\wedge}\right) {,}\quad \mathbf{P} \leftarrow \Delta \boldsymbol{\Theta} \cdot \mathbf{P}
        \label{equation 13}
    \end{equation}
the final estimate $\boldsymbol{\Theta}_\text{est}$ is the combination of all incremental estimates computed in the iterative loop.

\subsection{Learning Registration with Neural SDF}
\label{ssec:3.4}
\noindent\textbf{Learning SDF of Point Clouds.}
The implicit neural representation are commonly represented by multi-layer perceptrons (MLPs), to regress the continuous signed distance function (SDF) from a set of sampled points.
We learn the continuous SDF of the target point cloud $\textbf{Q}$ using implicit neural representation, which is denoted as $\Phi$.  
    \begin{equation}
        \operatorname{SDF}\left(\mathbf{x}\right) \approx \Phi\left(\mathbf{x} \ \lvert\ \textbf{Q} \right)
        \label{equation 15}
    \end{equation}
    
To train the networks, we sample a set of query points from space, denoted as \textbf{X}. The point closest to the query point $\mathbf{x}$ on the object surface is represented as $\mathbf{t}$. The predicted signed distance value is utilized to estimate $\hat{\mathbf{t}}$, the calculation method for $\hat{\mathbf{t}}$ refers to  Sec.1.1 of the appendix. We utilize a self-supervised loss \cite{GenSDF} to minimize the distance between $\hat{\mathbf{t}}$ and $\mathbf{t}$. The reconstructed self-supervised loss function is as follows,
    \begin{equation}
        \mathcal{L}_\text{rec}=\frac{1}{|\textbf{X}|} \sum_{j \in |\textbf{X}|}\left\|\hat{\mathbf{t}}_{j}-\mathbf{t}_{j}\right\|_{2}^{2}+\lambda_{q} \frac{1}{M} \sum_{q \in \textbf{Q}}\left\|\Phi\left(\mathbf{q}_{i}\ \lvert\ \textbf{Q}\right)\right\|_{1}
        \label{equation 18}
    \end{equation}
where $\lambda_{q}$ determines the weight of point cloud prediction and the $\ell_{1}$ loss is to constrain the SDF of target point clouds to a distance value of $0$.

Furthermore, we introduce an additional Eikonal term\cite{gropimplicit} to each query point $\mathbf{x}$ to regularize the $\Phi$ of SDF as follows.
    \begin{equation}
    \mathcal{L}_{\text {eikonal }}=\frac{1}{|\textbf{X}|}\sum_{j \in |\textbf{X}|}\left(\left\|\nabla \Phi\left(\mathbf{x}_{j}\right)\right\|_{2}-1\right)^{2}
        \label{equation 19}
    \end{equation}

\noindent\textbf{Learning Coarse-to-Fine Registration.} 
The learning strategy of the most SDF models only uses fixed query points in space and thus is not tied to the registration process. In addition, this method cannot focus on the query points in a certain area and change the query points adaptively, so it cannot learn accurate SDFs in local areas to further improve registration accuracy.

We first optimize the implicit SDF function. Instead of using fixed query points \textbf{X}, we use the points of the source point cloud $\textbf{P}$ as query points to optimize the SDF, allowing the model to learn the distances of farther points in the beginning. Then we perform one iteration to register the source point cloud and the target point cloud using \cref{equation 13} and add the new updated point cloud into query points.  These two steps are then alternately optimized. As the registration proceeds, the reconstructed loss function $\mathcal{L}_\text{rec}$ for learning SDF is also changed as follows. 
    \begin{equation}
        \mathcal{L}_\text{c2f}=\frac{1}{K} \sum_{k \in K}\left\|\hat{\mathbf{t}}_{t}-\mathbf{t}_{t}\right\|_{2}^{2}+\lambda_{q} \frac{1}{M} \sum_{q \in \textbf{Q}}\left\|\Phi\left(\mathbf{q}_{i}\ \lvert\ \textbf{Q}\right)\right\|_{1}
        \label{equation 18-1}
    \end{equation}
where $K$ is the number of source point clouds times the number of iterations: $K=N \ast  itrs$. It is noted that in the strategy, the query points are not sampled from the space. 

In summary, our loss function is defined as follows,
    \begin{equation}
        \mathcal{L}_{\text {reg }}=\mathcal{L}_{\text {c2f}}+\lambda_{e} \mathcal{L}_{\text { eikonal }}
        \label{equation 20}
    \end{equation}
where $\lambda_{e}=0.001$ is used to control the strength of the regularization of the Eikonal loss function.

The proposed coarse-to-fine registration strategy allows the model to dynamically acquire query points at various distances and focus the model's attention on learning the SDF values of query points relevant to the registration task during the iterative registration. This strategy will greatly enhance the performance of registration.

\section{Experiment}

\renewcommand{\arraystretch}{1} 
\begin{table*}[h]\setcounter{table}{0}
  \footnotesize
  \centering
  \setlength{\tabcolsep}{4.5mm}{
  \begin{threeparttable}
  \caption{ModelNet40: Performance on clean data and Gaussian noise data on unseen categories.}
    \label{tab1}
    \begin{tabular}{lcccccccc}
    \hline
    \multicolumn{1}{l}{\multirow{3}{*}{Methods}} & \multicolumn{4}{c}{Clean Data} & \multicolumn{4}{c}{Noise Data} \\
    \cmidrule(lr){2-5} \cmidrule(lr){6-9}
    \multicolumn{1}{c}{}  & \multicolumn{2}{c}{Rotation[deg] $\downarrow$}  & \multicolumn{2}{c}{Translation[m] $\downarrow$}  & \multicolumn{2}{c}{Rotation[deg] $\downarrow$} & \multicolumn{2}{c}{Translation[m] $\downarrow$}  \\
    \cmidrule(lr){2-3} \cmidrule(lr){4-5} \cmidrule(lr){6-7} \cmidrule(lr){8-9}
    & MAE  & RMSE & MAE     & RMSE    & MAE  & RMSE  & MAE  & RMSE\cr
    \hline
        ICP \cite{ICP}     & 13.887  & 23.751  & 0.0519 & 0.5951    &13.756 &23.459 &0.0511 &0.0585\cr
        PointNetLK\cite{PointNetLK} & 5.2969  & 18.123 & 0.0039 & 0.0241   &13.379 &27.531 &0.0392 &0.0376 \cr
        DCP-V2\cite{DCP}   & 2.0072  & 3.5501  & 0.0037  & 0.0050   &5.6346 &12.850 &0.0253 &0.0381 \cr
        FMR\cite{FMR}  & 4.9820 & 14.583 & 0.0067 & 0.0335   &7.2228 &16.658 &0.0233 &0.0337  \cr
        DeepGMR\cite{DeepGMR} & 9.3220  & 18.890  & 0.0559 & 0.0870   &8.5780 &17.693 &0.0531 &0.0849   \cr
        IFR\cite{ifr} & \textbf {1.1474}  & 5.0707  & 0.0100  & 0.0554  & 4.9289      & 11.335     & 0.0788       & 0.1271\cr
    \hline
        Ours       & 2.5886    & \textbf{3.1540}    & \textbf{0.0033}     &  \textbf{0.0048} & \textbf{3.9754}   & \textbf{9.1780}     & \textbf{0.0058}     & \textbf{0.0089}  \cr
    \hline
    \end{tabular}
    \end{threeparttable}
    }
    \vspace{-1.0em}
\end{table*}
\renewcommand{\arraystretch}{1.0} 
\begin{table*}[h]\setcounter{table}{1}
  \footnotesize
  \centering
  \setlength{\tabcolsep}{4.5mm}{
  \begin{threeparttable}
  \caption{ModelNet40: Performance on partial point clouds or on the density changes.}
  \label{tab2}
    \begin{tabular}{lcccccccc}
    \hline
    \multicolumn{1}{l}{\multirow{3}{*}{Methods}} & \multicolumn{4}{c}{Partial Point Clouds} & \multicolumn{4}{c}{Density Changes} \\
    \cmidrule(lr){2-5} \cmidrule(lr){6-9}
    \multicolumn{1}{c}{}  & \multicolumn{2}{c}{Rotation[deg] $\downarrow$}  & \multicolumn{2}{c}{Translation[m] $\downarrow$}  & \multicolumn{2}{c}{Rotation[deg] $\downarrow$} & \multicolumn{2}{c}{Translation[m] $\downarrow$}  \\
    \cmidrule(lr){2-3} \cmidrule(lr){4-5} \cmidrule(lr){6-7} \cmidrule(lr){8-9}
    & MAE  & RMSE & MAE      & RMSE    & MAE  & RMSE  & MAE  & RMSE\cr
    \hline
        ICP \cite{ICP} & 23.761    & 31.645   & 0.1168    & 0.1279
        & 14.435    & 23.976   & 0.0542    & 0.0612 \cr
        PointNetLK\cite{PointNetLK} & 81.365    & 79.634   & 0.1236    & 0.1925   
        & 18.693    & 30.121   & 0.0446    & 0.0408 \cr
        DCP-V2\cite{DCP} & 26.927    & 43.275   & 0.0981     & 0.1168   
        & 39.617    & 59.627   & 0.0091     & 0.0318 \cr
        FMR\cite{FMR} & 24.813    & 30.677   & 0.1342      & 0.1511      
        & 7.5764   & 15.830   & 0.0201      & 0.0208 \cr
        DeepGMR\cite{DeepGMR} & 10.227      & 19.720     & 0.0596       & 0.0915   
        &-    &-    &-    &-   \cr
        IFR\cite{ifr}  & 5.0400      & 11.728     & 0.0232       & 0.0233  
        & \textbf{3.4699}      & 9.3591     & 0.0395       & 0.1122 \cr
    \hline
        Ours & \textbf{4.1362}    & \textbf{11.596}   & \textbf{0.0056}     & \textbf{0.0103}     
        & 3.8191    & \textbf{8.839}   & \textbf{0.0084}     & \textbf{0.0098} \cr
    \hline
    \end{tabular}
    \end{threeparttable}
    }
    \vspace{-1.0em}
\end{table*}
We conduct experiments using the ModelNet40\cite{ModelNet} datasets, containing 12311 CAD models across 40 classes, split into 20 for training and 20 for cross-class testing. Additionally, we utilize 3DMatch\cite{3dmatch} as a benchmark, comprising real-world indoor datasets captured via Kinect cameras, and Stanford 3D scanning dataset\cite{stanford} for evaluating our method's performance across various real-world object scans.
\subsection{Clean Point Cloud}
\label{ssec:4.3}
We follow the experimental setup of PointNetLK\cite{PointNetLK}, splitting the ModelNet40 dataset into two parts, with 20 categories for training and testing. We uniformly sample 1024 points from the point cloud data and normalize to a cell box at the origin $[0,1]^{3}$, then uniformly sample in $\left[0,45^{\circ}\right]$ along the rotation of each axis, and translate in $[-0.5, 0.5]$. From the \cref{tab1}, it can be seen that ICP\cite{ICP}, PointNetLK\cite{PointNetLK}, FMR\cite{FMR} and DeepGMR\cite{DeepGMR} still have large errors. The effect of our method is comparable to IFR\cite{ifr} and DCP\cite{DCP}.


\begin{figure}[ht]
    \centering
    \includegraphics[width=0.5\textwidth]{"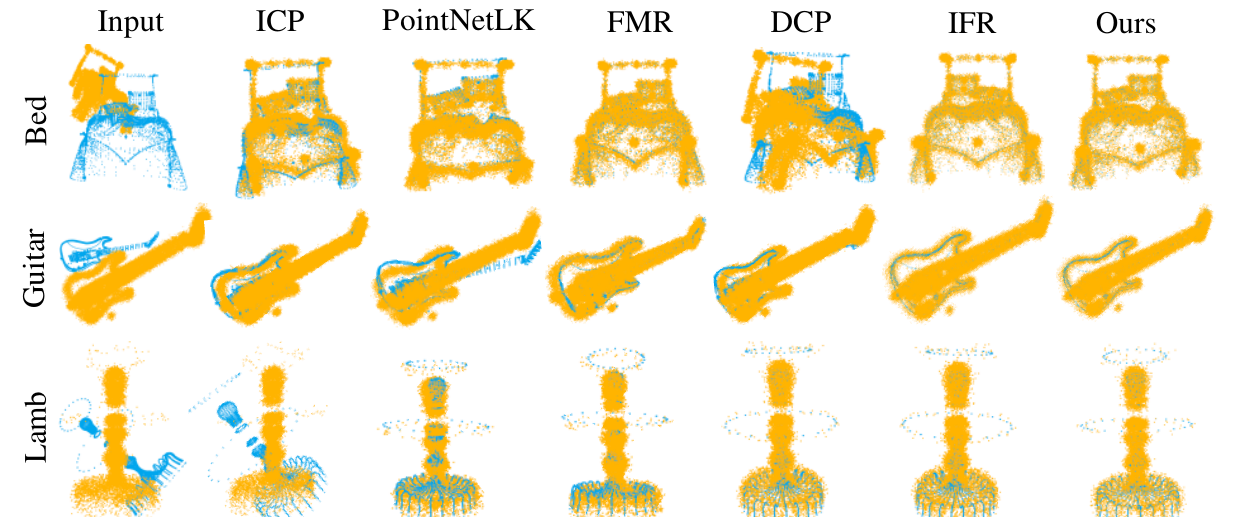"}
     \caption{Qualitative evaluation on Gaussian noise point clouds.}
    \label{fig4}
    \vspace{-1.0em}
\end{figure}

\begin{figure}[ht]
    \centering
    \includegraphics[width=0.5\textwidth]{"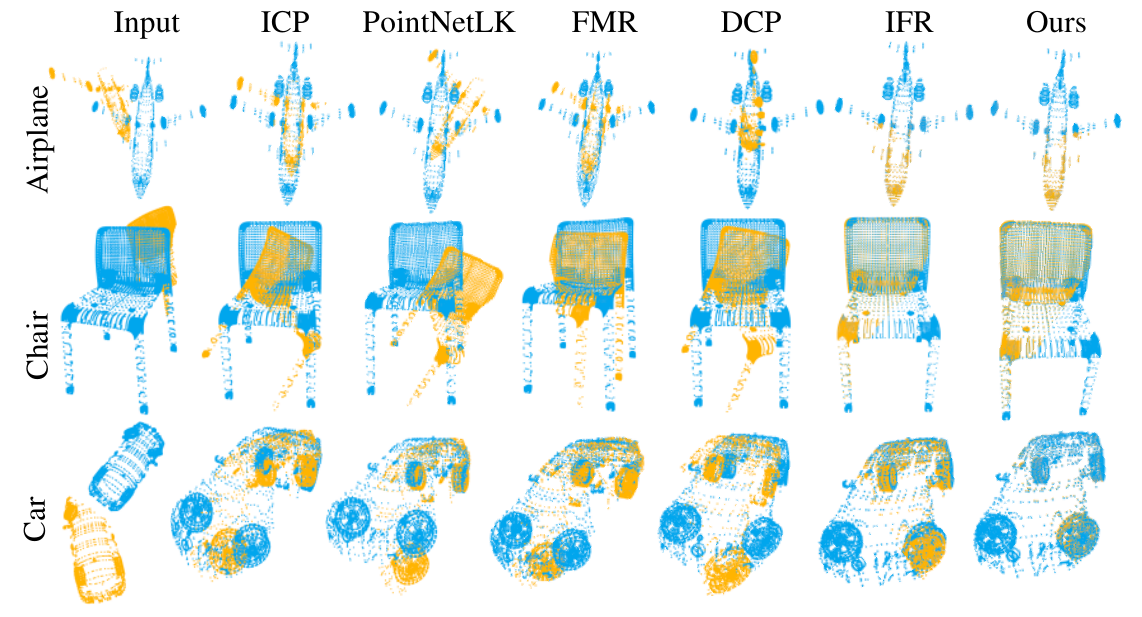"}
    \vspace{-1.0em}
    \caption{Qualitative evaluation on partial point clouds.}
    \label{fig5}
    \vspace{-1.0em}
\end{figure}

\begin{figure}[h]
    \centering
    \includegraphics[width=0.5\textwidth]{"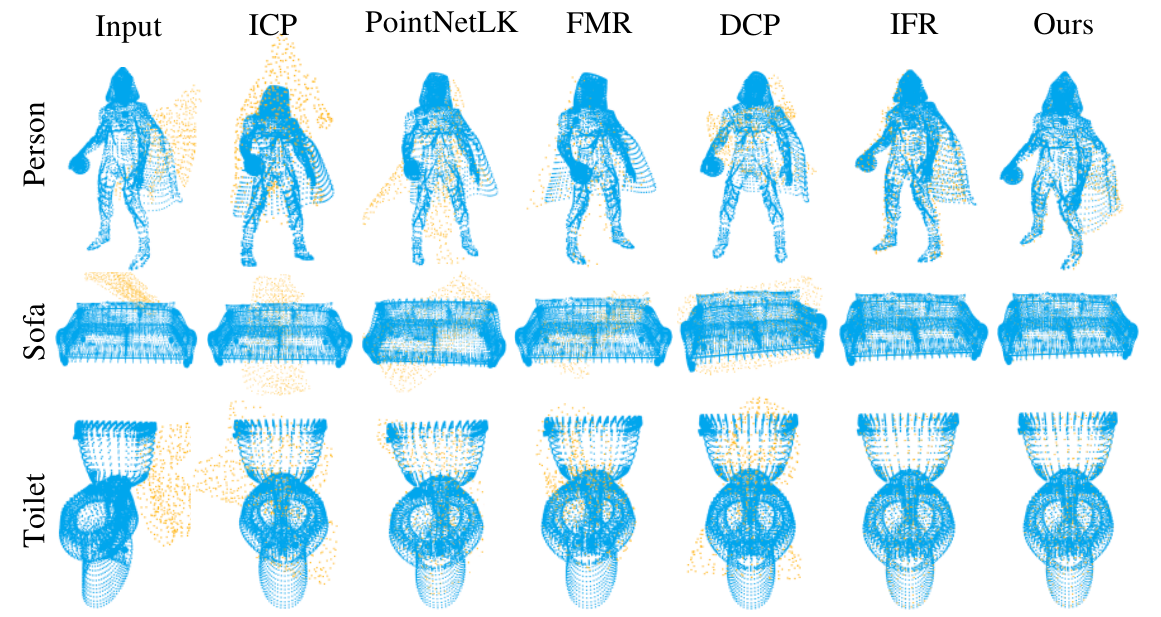"}
    \vspace{-2.0em}
    \caption{Qualitative evaluation on density changes.}
    \label{fig6}
    \vspace{-1.0em}
\end{figure}

\subsection{Gaussian Noise}

In order to discuss the robustness of SDFReg to Gaussian noise on point clouds, we independently added Gaussian noise sampled from $\mathcal{N}\left(0,0.02^{2}\right)$ to each point coordinate of the clean target point cloud in the ModelNet40 dataset. In this experiment, the SDFReg model is also trained on clean data, and the experimental configuration is the same as in \cref{ssec:4.3}. It can be seen from the \cref{tab1} that ICP\cite{ICP}, DCP\cite{DCP}, and PointNetLK\cite{PointNetLK} are greatly affected by noise. 
While FMR\cite{FMR} and IFR\cite{ifr} can mitigate the impact of noise on point cloud registration, our method surpasses them in overall registration quality. This is attributed to the effectiveness of the signed distance function in fitting noisy data, making our method outperform others in robustness experiments with noisy point clouds.
Qualitative results are shown in \cref{fig4}.


\subsection{Partial Visibility}
We evaluated the performance on partial point clouds in the ModelNet40 dataset, where two point clouds do not completely overlap in extent. Aligning partially visible point clouds is a common registration scenario. 
In the real world, the template is usually a complete 3D model, and the source is a 2.5D scan. During the test, for the source point cloud, we separately create a random plane passing through the origin, translate it along the normal, and retain $70\%$ of the points. The results are shown in \cref{tab2}. The Quantitative experimental results indicate that all competitors including ICP\cite{ICP}, PointNetLK\cite{PointNetLK}, DCP\cite{DCP}, FMR\cite{FMR}, DeepGMR\cite{DeepGMR}, and IFR\cite{ifr} are not as good as our method in partial point clouds registration. The qualitative results are shown in \cref{fig5}.

\subsection{Density Changes}
To demonstrate that our SDFReg model can handle significant density differences, we conducted this experiment on the ModelNet40 dataset. Referring to FMR\cite{FMR}, we input two point clouds, each with 10,000 points, and randomly removed 95\% of the points from one of the original point clouds (source), resulting in a 20x density difference. \Cref{tab2} shows that ICP\cite{ICP}, PointNetLK\cite{PointNetLK}, and DCP\cite{DCP} have difficulty in aligning these point clouds with large density differences. Note that DeepGMR\cite{DeepGMR} is unable to handle the case of point cloud density due to significant difference. Our method can effectively align these point clouds and yield comparable results to IFR\cite{ifr}. The qualitative results are shown in \cref{fig6}.

\subsection{Robustness Test}
To further demonstrate the robustness of each method in terms of rotation and translation estimation, we calculated the standard deviation and average registration accuracy under different situations with clean, partial, noise, and density-varying point clouds. The experimental results prove that our method is minimally affected by the quality of the input point cloud during registration, indicating that our approach exhibits the best robustness.
\renewcommand{\arraystretch}{1.0} 
\begin{table}[ht]
\vspace{-1.0em}
  \footnotesize
  \centering
  \setlength{\tabcolsep}{2.2mm}{
  \begin{threeparttable}
  \caption{Quantitative comparison in robustness testing.}
  \label{tab3}
    \begin{tabular}{l@{}|cccc}
    \hline
    \diagbox[width=60pt, height=20pt]{Methods}{Robust}  & SD(R) $\downarrow$  & SD(t) $\downarrow$ & AVG(R) $\downarrow$  & AVG(t) $\downarrow$     \cr
    \hline
        ICP \cite{ICP}        & 4.886  & 0.032  & 16.46     & 0.069\cr
        PointNetLK \cite{PointNetLK}\; & 34.89  & 0.053  & 29.68  & 0.051 \cr
        DCP-V2 \cite{DCP}      &17.84  & 0.044  & 18.55   & 0.034    \cr
        FMR \cite{DeepGMR}      & 9.182   & 0.059   & 11.14    & 0.046  \cr
        DeepGMR \cite{ifr}    & 8.258  & 0.033  & 9.376   & 0.056   \cr
        IFR \cite{ifr}        & 1.814  & 0.042  & 3.647   & 0.035  \cr
    \hline
        Ours      & \textbf{0.706}     & \textbf{0.002}    & \textbf{3.321}     &  \textbf{0.005} \cr
    \hline
    \end{tabular}
    \end{threeparttable}
    }
    \vspace{-1.0em}
\end{table}

\subsection{Evaluation on 3DMatch}
We evaluated our method on the 3DMatch dataset, which is a collection of several real-world indoor datasets, including complex 3D RGB-D scans obtained from multiple scenes such as offices, hotels, and kitchens. The transformed source point cloud is taken as the target point cloud, and the transformation is also generated by random sampling. The rotation is initialized in the range of [0, 60°] and translation is initialized in the range of [0, 1.0]. As shown in \cref{tab3}, our method is more effective than all comparative methods.
\renewcommand{\arraystretch}{1} 
\begin{table}[ht]
\vspace{-1.0em}
  \footnotesize
  \centering
  \setlength{\tabcolsep}{3.5mm}{
  \begin{threeparttable}
  \caption{Quantitative comparisons on the 3DMatch dataset.}
  \label{tab3}
    \begin{tabular}{lcccc}
    \hline
    \multirow{2}{*}{Methods}&
    \multicolumn{2}{c}{Rotation[deg] $\downarrow$}&\multicolumn{2}{c}{Translation[m] $\downarrow$} \cr
    \cmidrule(lr){2-3} \cmidrule(lr){4-5}
    & RMSE      & Median     & RMSE       & Median \cr
    \hline
        ICP \cite{ICP}        & 24.772   & 4.501   & 1.064     & 0.149 \cr
        PointNetLK \cite{PointNetLK} & 28.894  & 7.596   & 1.098      & 0.260  \cr
        DCP-V2 \cite{DCP}     & 53.903    & 23.659    & 1.823     & 0.784   \cr
        DeepGMR \cite{DeepGMR}    & 32.729      & 16.548      & 2.112       & 0.764  \cr
        IFR \cite{ifr}    & 23.814   & 4.191    & 1.057       & 0.161   \cr
    \hline
        Ours      & \textbf{15.623}     & \textbf{3.1562}    & \textbf{0.065}     &  \textbf{0.013} \cr
    \hline
    \end{tabular}
    \end{threeparttable}
    }
    \vspace{-1em}
\end{table}


\begin{table}[ht]
    \footnotesize
    \centering
    \setlength{\tabcolsep}{1.7mm}{
    \caption{Ablation study for using coarse-to-fine learning strategy. The definition of each loss function is shown in the appendix.}
    \label{tab4}
    \begin{tabular}{lcccccc}
      \toprule
        Method         & MAE(R) $\downarrow$   & RMSE(R) $\downarrow$  & MAE(t) $\downarrow$    & RMSE(t) $\downarrow$   \\
      \midrule
    $\mathcal{L}_{\text {self}}$    & 6.8108    & 19.513   & 0.0154    & 0.0518  \\
    $\mathcal{L}_{\text {c2f}}$    & 4.1376    & 11.681   & 0.0062    & 0.0106  \\
    $\mathcal{L}_{\text {self}} + \mathcal{L}_{\text {eikonal}}$    & 5.3418    & 16.263   & 0.0098    & 0.0186  \\
    $\mathcal{L}_{\text {c2f}} + \mathcal{L}_{\text {eikonal}}$     & \textbf{2.5887}    & \textbf{8.5548}   & \textbf{0.0034}    & \textbf{0.0049}  \\

      \bottomrule
    \end{tabular}}
    \vspace{-1.0em}
\end{table}

\subsection{Ablation Study}
In this section, we analyze the effects of each improvement represented in the loss function. All experiments are conducted in the same environment as the experiments in \cref{ssec:4.3}.  We show the quantitative results on the ModelnNet40 dataset in \cref{tab4}. It can be observed from the table that the results using the coarse-to-fine strategy $\mathcal{L}_{\text{c2f}}$ are largely superior to the ones based on traditional self-supervised loss $\mathcal{L}_{\text{loss}}$. Eikonal loss $\mathcal{L}_{\text{eikonal}}$ will further promote the effectiveness of the model in point cloud registration. After merging the two losses, the model's registration performance significantly improves. 
\section{Conclusion}    
In this paper, we propose SDFReg, a novel correspondence-free point cloud registration method, which introduces a neural implicit surface to represent the target point cloud. By optimizing the distance metric between the source point cloud and the implicit surface, the accurate pose estimation is finally calculated. We conduct experiments on synthetic and real data, demonstrating the network's ability to handle various input imperfections. We found that registering point cloud and neural implicit function make the model more robust to point density, noise, and missing data. 
\vspace{-1em}

\small
\bibliographystyle{IEEEbib}
\bibliography{reference}

\end{document}


\sloppy

\def\x{{\mathbf x}}
\def\L{{\cal L}}

\title{supplementary Material for SDFReg}
%
\name{Leida Zhang, Zhengda Lu, Kai Liu, Yiqun Wang\textsuperscript{\Envelope}}
\address{}

\maketitle

\section{Appendix}
\subsection{Training Strategy}
\label{ssec:3.4}
The key of our SDFReg is to enable the model not only to represent the point cloud well as the implicit neural surface, but also to calculate the exact distance from the query point to the implicit surface for registration. 
For this purpose, we design a coarse-to-fine registration strategy to alternately optimize the implicit function fitting and the registration.

\noindent\textbf{Learning SDF of Point Clouds.} We use a semi-supervised learning method \cite{GenSDF}, to learn a continuous SDF for the target point cloud $\textbf{Q}$. We choose this method because it can directly infer the point cloud's SDF through the input of point cloud $\textbf{Q}$ in an effective way, so there is no need to retrain the SDF from scratch
for the shapes of unknown categories. 
Registration tasks often require inferring poses for a large number of shapes of unknown categories, making this method a highly efficient choice for such tasks.

The model is trained by a self-supervised loss function. We sample a set of query points from space, denoted as \textbf{X}. For each query point $ \mathbf{x} \in \mathbb{R}^{3}$, we approximate the projection of the query point on the surface by using its nearest point $\mathbf{q} \in \textbf{Q}$. To achieve self-supervised learning of SDF values, the projection position is defined as follows.
    \begin{equation}
        \mathbf{t}=\underset{\mathbf{q} \in \textbf{Q}}{\arg \min }\|\mathbf{x}-\mathbf{q}\|_{2}
        \label{equation 16}
    \end{equation}
Given the SDF function $\Phi$, we use
    \begin{equation}
        \hat{\mathbf{t}}=\left\{\begin{array}{ll}
        \mathbf{x}-\frac{\mathbf{x}-\mathbf{t}}{\|\mathbf{x}-\mathbf{t}\|} \Phi(\mathbf{x}\ \lvert\ \textbf{Q}) & \text { if } \Phi(\mathbf{x}\ \lvert\ \textbf{Q}) \geq 0, \\\\
        \mathbf{x}+\frac{\mathbf{x}-\mathbf{t}}{\|\mathbf{x}-\mathbf{t}\|} \Phi(\mathbf{x}\ \lvert\ \textbf{Q}) & \text { if } \Phi(\mathbf{x}\ \lvert\ \textbf{Q}) < 0
        \end{array}\right.
        \label{equation 17}
    \end{equation}
to approximate the point $\mathbf{t}$ on the object's surface closest to the query point $\mathbf{x}$. The predicted signed distance value is utilized to estimate $\hat{\mathbf{t}}$. The objective of the self-supervised loss is to minimize the distance between $\hat{\mathbf{t}}$ and $\mathbf{t}$. The $\ell_{1}$ loss is to constrain the SDF of target point clouds to a distance value of $0$.  The loss function is as follows,
    \begin{equation}
        \mathcal{L}_\text{self}=\frac{1}{|\textbf{X}|} \sum_{j \in |\textbf{X}|}\left\|\hat{\mathbf{t}}_{j}-\mathbf{t}_{j}\right\|_{2}^{2}+\lambda_{q} \frac{1}{M} \sum_{q \in \textbf{Q}}\left\|\Phi\left(\mathbf{q}_{i}\ \lvert\ \textbf{Q}\right)\right\|_{1}
        \label{equation 18}
    \end{equation}
where $\lambda_{q}=0.01$ determines the weight of point cloud prediction, $|\textbf{X}|$ represents the number of query point sets $\textbf{X}$.

\subsection{Network Architecture}
\label{ssec:4.h}
\begin{figure*}
  \centering
   \includegraphics[width=1.0\linewidth]{"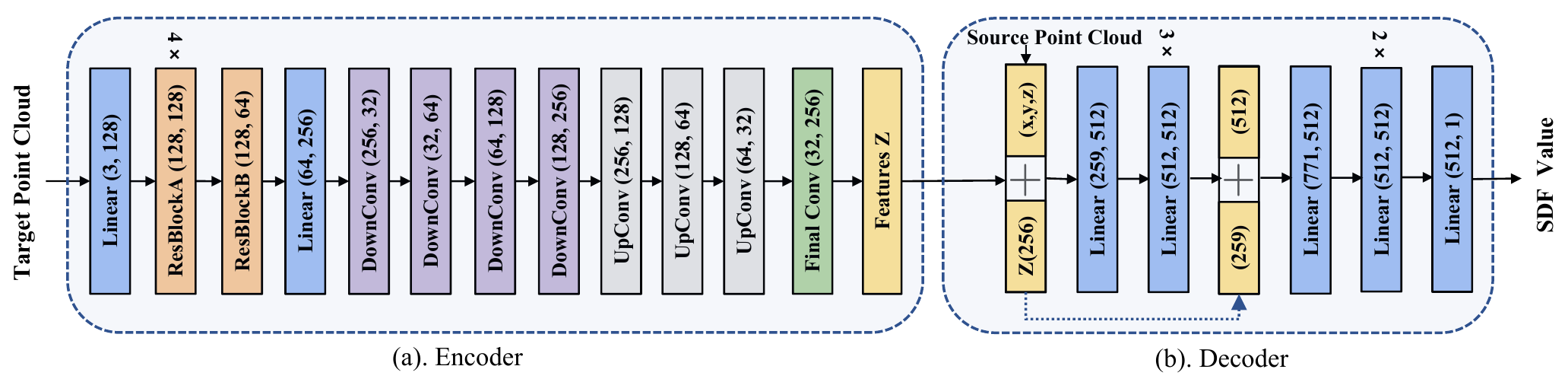"}
   \caption{Network Architectures. The encoder takes the target point cloud as the input and output feature $z$ and the decoder takes the coordinate of the source point cloud as the input and output SDF value.}
   \label{figm}
\end{figure*}

As shown in \Fref{figm}, our network architecture is a decoder-encoder architecture, which is widely used in many literature~\cite{IM-Net,deepsdf,OccNet,Conv_occupancy}. Specifically, we use the architecture \cite{GenSDF} as the backbone of our model. The encoder takes the target point cloud as input and uses an improved PointNet \cite{pointnet, OccNet} with 5 ResNet blocks to extract geometric features. A series of down-sampling and up-sampling convolutions with skip connections are also employed to aggregate spatial information and integrate both local and global information, similar to UNets\cite{unet}. The size of the output feature $z$ of encoder is set to 256. Similar to the work \cite{GenSDF}, our decoder takes the coordinate of the query point in the \textit{source point cloud} and the feature $z$ as input and utilizes 8-layer MLPs with 512 hidden dimensions to predict the SDF value. To reduce the information loss of the feature, a skip connection is also applied, as indicated by the dotted arrow in the figure. The complete model architecture is illustrated in \cref{figm}.

\section{Experiment}
\subsection{Implentation details}
Our model needs to register the point cloud in the loop iteration, when the relative distance error during the iteration is less than a given threshold, the iteration ends. The relative distance error is defined as $\epsilon = \mid \operatorname{D}_{n}-\operatorname{D}_{n-1} \mid/\operatorname{D}_{n}$, where $\operatorname{D}_{n}$ refers to the SDF value between the source point cloud and the target point cloud when iterating to the $n$ steps. The algorithm stops when  $\epsilon <0.001$ or $n > 30$ in our implementation. All experiments are implemented on a single NVIDIA GeForce RTX 3090 GPU (24 GB memory).

\subsection{Evaluation Metrics}
Following the FMR\cite{FMR}, DCP\cite{DCP} and IFR\cite{ifr}, we evaluate the registration model by calculating the error between the estimated rotation and translation and the ground truth, the rotation matrix $\mathbf{R}$ and translation vector $\mathbf{t}$ are obtained from the transformation parameters $\mathbf{\Theta}$. Therefore, the rotation and translation errors are calculated by:
   \begin{equation}
    \begin{cases}
       \operatorname{Error}(\mathbf{R})=\arccos ((\operatorname{trace}(\mathbf{R}_{gt}^{-1}   \hat{\mathbf{R}}) + 1)/2)\frac{180}{\pi}, \\  \operatorname{Error}(\mathbf{t})=\left\|\mathbf{t}_{g t}-\hat{\mathbf{t}}\right\|_{2}
       \label{equation 21}
    \end{cases}
   \end{equation}
where, $\left\{\mathbf{R}_{g t}, \mathbf{t}_{g t}\right\}$ and $\{\hat{\mathbf{R}}, \hat{\mathbf{t}}\}$ represent the ground truth and predicted transformations, respectively.

For consistency with previous work\cite{DCP}, we compute the root mean squared error (RMSE), and mean absolute error (MAE) between the ground truth and predicted values. Ideally, all these error metrics should be zero if rigid alignment is perfect. All angle measurements in the results are in degrees.

\subsection{Datasets}

\noindent\textbf{ModelNet40:} We train and test our method on the ModelNet40\cite{ModelNet} dataset, which consists of 12311 CAD models from 40 object classes. We follow the experimental setup of PointNetLK\cite{PointNetLK} by dividing the dataset into two parts: 20 classes for training, and another 20 classes for cross-class testing. The source point cloud is the vertices from ModelNet40, which we uniformly sample 1024 points and normalize to a unit box at the origin $[0, 1]^{3}$. Following\cite{DCP}, the rotation along each axis is uniformly sampled in $\left[0, 45^{\circ}\right]$, and the translation is in $[-0.5, 0.5]$. The source point cloud is rigidly transformed to generate the target point cloud, which is then used to train and evaluate the registration model.

\noindent\textbf{3DMatch:} 3DMatch\cite{3dmatch} is a collection of several real-world indoor datasets and a widely used registration benchmark that captures data using Kinect cameras, including complex 3D RGB-D scans obtained from multiple scenes such as offices, hotels, and kitchens. Following PointNetLK-Revisited\cite{pointlkre}, we chose 8 categories of scenes from 7-Scenes\cite{7scene} and SUN3D\cite{sun3d} for testing, and the rest are used for training.

\noindent\textbf{Stanford 3D Scan:} We also use Stanford 3D scanning dataset\cite{stanford,stanford1} to test the performance of our method. This dataset contains a collection of 3D scans depicting various real-world objects, which allows for better testing of the generalization of our models.

\subsection{Qualitative Results}
\begin{figure}[h]
  \centering
   \includegraphics[width=0.9\linewidth]{"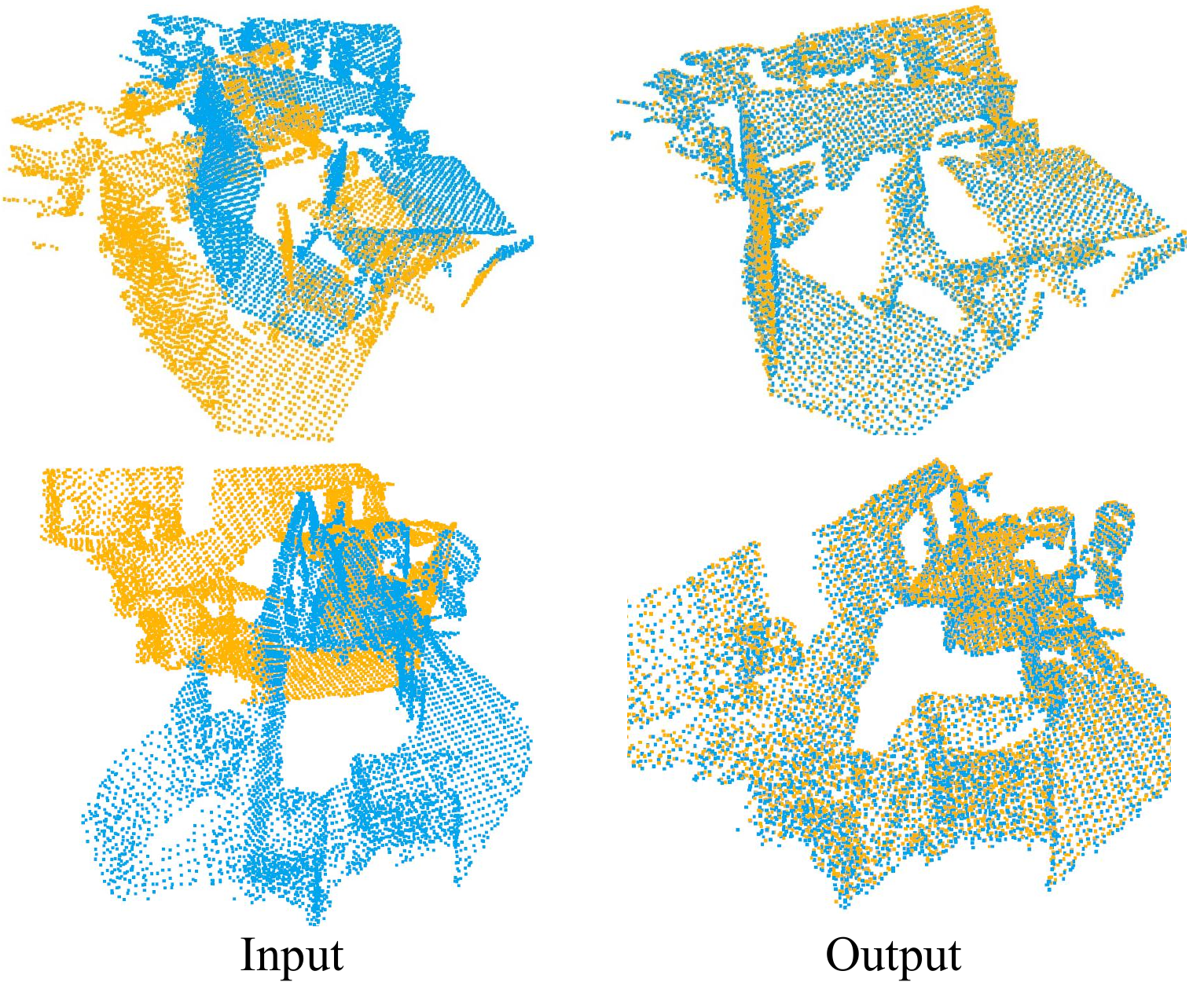"}
    \vspace{-1.0em}
   \caption{Example results of visualization on the dataset of 3DMatch.}
   \label{fig7}
   \vspace{-1.0em}
\end{figure}

\begin{figure}[h]
\vspace{-1.0em}
  \centering
   \includegraphics[width=0.95\linewidth]{"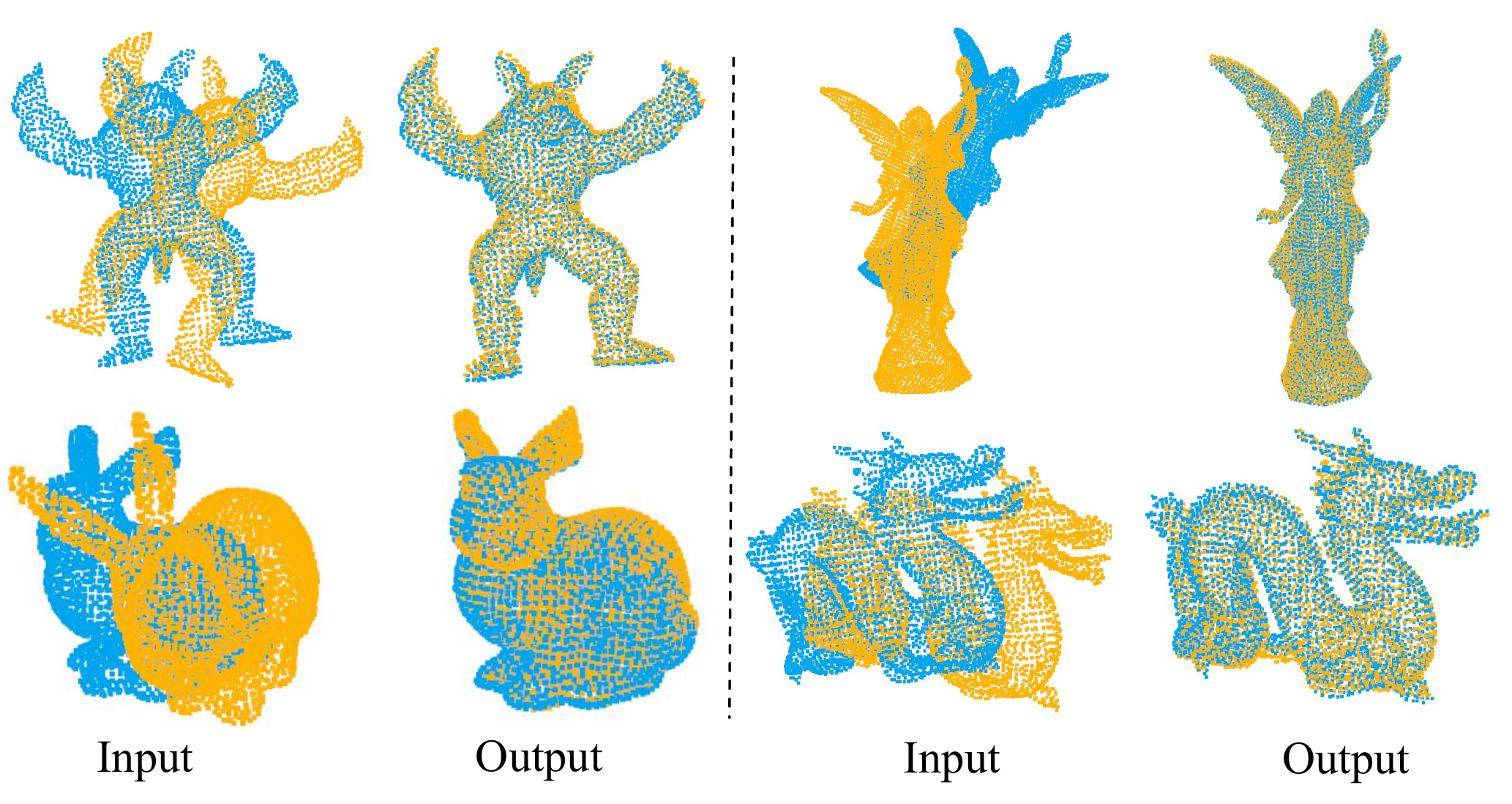"}
   \vspace{-1.0em}
   \caption{Example results of visualization on on Stanford 3D Scan.}
   \label{fig8}
\end{figure}

To verify the registration ability of the model on real scanning datasets, we visualize its performance on the 3DMatch dataset in \Fref{fig7}. Furthermore, we tested the model trained on ModeNet on the Stanford 3D scan dataset, and the results are shown in \Fref{fig8}. Our model demonstrates good generalization ability in real-world scenes.

\small
\bibliographystyle{IEEEbib}
\bibliography{reference}